\documentclass{article} 
\usepackage[final]{templates/colm/colm2026_conference}

\usepackage{microtype}
\usepackage{hyperref}
\usepackage{url}
\usepackage{booktabs}
\usepackage{lineno}

\usepackage{graphicx}
\usepackage{amsmath}
\usepackage{amssymb}
\usepackage{xcolor}
\usepackage{multirow}
\usepackage{pifont}
\usepackage{tikz}
\usepackage{pgfplots}
\pgfplotsset{compat=1.18}
\usetikzlibrary{patterns,positioning}

\definecolor{darkblue}{rgb}{0, 0, 0.5}
\hypersetup{colorlinks=true, citecolor=darkblue, linkcolor=darkblue, urlcolor=darkblue}

\makeatletter
\newcommand{\loadcachedbibcites}{%
  \@ifundefined{b@radford2021clip}{\bibcite{alayrac2022flamingo}{{1}{2022}{{Alayrac et~al.}}{{Alayrac, Donahue, Luc, Miech, Barr, Hasson, Lenc, Mensch, Millican, Reynolds, Ring, et~al.}}}
\bibcite{beyer2024paligemma}{{2}{2024}{{Beyer et~al.}}{{Beyer, Steiner, Pinto, et~al.}}}
\bibcite{chaudhury2025chameleonbench}{{3}{2025}{{Chaudhury \& Shiromani}}{{Chaudhury and Shiromani}}}
\bibcite{dai2023instructblip}{{4}{2023}{{Dai et~al.}}{{Dai, Li, Dong, Tiong, Li, Savarese, and Hoi}}}
\bibcite{fu2023mme}{{5}{2023}{{Fu et~al.}}{{Fu, Chen, Shen, Qin, Zhang, Lin, Yang, Zheng, Li, Sun, Wu, Ji, Shan, and He}}}
\bibcite{jain2019attention}{{6}{2019}{{Jain \& Wallace}}{{Jain and Wallace}}}
\bibcite{li2023seedbench}{{7}{2023{a}}{{Li et~al.}}{{Li, Wang, Wang, Ge, Ge, and Shan}}}
\bibcite{li2022blip}{{8}{2022}{{Li et~al.}}{{Li, Li, Xiong, and Hoi}}}
\bibcite{li2023hallucination}{{9}{2023{b}}{{Li et~al.}}{{Li, Du, Zhou, Wang, Zhao, and Wen}}}
\bibcite{liu2023llava}{{10}{2023}{{Liu et~al.}}{{Liu, Li, Wu, and Lee}}}
\bibcite{liu2025seeing}{{11}{2025}{{Liu et~al.}}{{Liu, Chen, Wang, and Zhao}}}
\bibcite{long2025coe}{{12}{2025}{{Long et~al.}}{{Long, Oh, Park, and Li}}}
\bibcite{nostalgebraist2020logitlens}{{13}{2020}{{Nostalgebraist}}{{}}}
\bibcite{radford2021clip}{{14}{2021}{{Radford et~al.}}{{Radford, Kim, Hallacy, Ramesh, Goh, Agarwal, Sastry, Askell, Mishkin, Clark, Krueger, and Sutskever}}}
\bibcite{rohrbach2018object}{{15}{2018}{{Rohrbach et~al.}}{{Rohrbach, Hendricks, Burns, Darrell, and Saenko}}}
\bibcite{sahay2025compass}{{16}{2025}{{Sahay et~al.}}{{Sahay, Pandya, Nagale, Lin, Shiromani, Zhu, and Sunishchal}}}
\bibcite{shiromani2026hypocrisygap}{{17}{2026}{{Shiromani et~al.}}{{Shiromani, Chaudhury, and Kunda}}}
\bibcite{thomas2026promoralbench}{{18}{2026}{{Thomas et~al.}}{{Thomas, Shiromani, Chaudhry, Li, Sharma, Zhu, and Dev}}}
\bibcite{wang2024qwen2vl}{{19}{2024}{{Wang et~al.}}{{Wang, Li, et~al.}}}
\bibcite{wang2022self}{{20}{2022}{{Wang et~al.}}{{Wang, Wei, Schuurmans, Le, Chi, Narang, Chowdhery, and Zhou}}}
\bibcite{wiegreffe2019attention}{{21}{2019}{{Wiegreffe \& Pinter}}{{Wiegreffe and Pinter}}}
\bibcite{yu2023mmvet}{{22}{2023}{{Yu et~al.}}{{Yu, Yang, Li, Wang, Lin, Liu, Wang, and Wang}}}
\bibcite{zhou2023llavabench}{{23}{2023}{{Zhou et~al.}}{{Zhou, Fu, Chen, Liu, Lin, Yan, and Chen}}}}{}%
}
\makeatother

\title{Visuals Lie, Consistency Speaks: Disentangling Spatial Attention from Reliability in Vision-Language Models}

\author{
Logan Mann$^{2}$ \quad Yi Xia$^{1}$ \quad Ajit Saravanan$^{2}$ \quad Ishan Dave$^{3}$ \quad Saadullah Ismail$^{1}$ \\[3pt]
Shikhar Shiromani$^{4}$ \quad Emily Huang$^{1}$ \quad Ruizhe Li$^{1}$ \quad Kevin Zhu$^{1}$ \\[5pt]
$^{1}$University of California, Santa Barbara \quad $^{2}$Algoverse AI Research \\[2pt]
$^{3}$University of California, Berkeley \quad $^{4}$Independent Researcher
}

\begin{document}
\ifcolmsubmission
\linenumbers
\fi
\maketitle
\loadcachedbibcites

\begin{abstract}
Multimodal Foundation Models (MFMs) are rapidly evolving from simple pattern matchers into reasoning agents. As these systems are used in higher-stakes settings, reliability, or knowing when a model may hallucinate, becomes critical. A common intuition in the field, which we call the \textit{Attention-Confidence Assumption}, is that reliability follows from ``structural'' visual perception: if a model focuses tightly on relevant image regions, its subsequent answer should be trustworthy. Conversely, scattered attention is often assumed to signal confusion.

We challenge this assumption through \textit{VLM Reliability Probe (VRP)}, a systematic cross-family investigation into reliability signals in contemporary Vision-Language Models (VLMs). We introduce ``structural attention'' metrics, including cluster counts ($C_k$) and spatial entropy ($H_s$) to quantify the coherence of the visual encoder's gaze. To capture the dynamics of this gaze, we further track attention evolution ($\Delta H_s$) across all layers. This analysis reveals a critical ``Symbolic Detachment'': models often exhibit ``Early Locking'' of visual features only to diffuse attention in later layers, effectively severing the link between early perception and final generation. Contrary to the grounding hypothesis, our results demonstrate a ``Cluster Failure'': spatial attention patterns possess near-zero correlation (\textbf{$R \approx 0.001$}) with model accuracy. Instead, we find that reliability is fundamentally a phenomenon of \textit{generation dynamics} and internal state distributions. Self-Consistency (SC), the agreement rate across sampled reasoning paths, emerges as the dominant predictor of truth (\textbf{$R=0.429$}). By aggressively scaling causal interventions, we further demonstrate a massive architectural divergence: LLaVA ``locks'' its prediction in a fragile late-stage structural bottleneck, whereas PaliGemma and Qwen2-VL distribute reliability globally, showing extreme resilience even when $\sim 50\%$ or more of their most predictive layer is destroyed. These findings suggest that for current VLMs, reliability signals are detached from visual grounding maps, and are best inferred from generation-time dynamics and hidden-state probes.

\end{abstract}

\section{Introduction}
\label{sec:intro}

The integration of vision and language into Multimodal Foundation Models (MFMs) promises a future where AI agents can perceive and reason about the physical world. However, this promise is threatened by hallucination, the tendency of models to generate confident but factually incorrect assertions. To deploy these models in safety-critical domains (e.g., robotics, medical imaging), we must be able to quantify their reliability.

Traditionally, interpretability research has looked to the "Attention Mechanism" as a window into the model's mind \cite{jain2019attention}. In Vision-Language Models (VLMs), this manifests as the \textit{Attention-Confidence Assumption}: the belief that a model's reliability is correlated with the quality of its visual grounding. If a model is asked, "Is there a dog?" and it focuses sharply on the dog, we assume that it "knows" the answer. If its attention is diffuse or focuses on the background, we assume that it is hallucinating.

In this work, we rigorously test this assumption across three representative VLM families (LLaVA-1.5, PaliGemma, and Qwen2-VL) \cite{liu2023llava,beyer2024paligemma,wang2024qwen2vl}. We perform a comprehensive analysis of reliability signals by comparing "structural" metrics derived from visual cross-attention against "linguistic" metrics derived from generation dynamics. We explicitly position novelty at the hidden-state reliability probe and cross-family layer-wise analysis; attention-failure and self-consistency are treated as important prior findings that we extend and calibrate in the VLM setting.

\textbf{Terminology note.} We use \emph{VLM} as the default term throughout; \emph{MFM} and \emph{LVLM} are used only when matching prior-work phrasing.

\noindent\textbf{Reproducibility.} Code and evaluation scripts are available at \url{https://github.com/itsloganmann/VLM-Reliability-Probe} (prompts, split definitions, and probe training pipeline).

\section{Related Work}
\label{sec:related}
Large vision-language models (LVLMs) are built on foundation architectures such as CLIP-style image encoders and large language backbones, enabling strong instruction-following and open-ended reasoning \cite{radford2021clip,alayrac2022flamingo,li2022blip,liu2023llava,dai2023instructblip}. Reliability and grounding concerns emerge when these models generate fluent but incorrect outputs, which has motivated benchmark-centric studies of hallucination in captioning and VQA \cite{rohrbach2018object,li2023hallucination}. Beyond LLaVA-Bench \cite{zhou2023llavabench}, recent evaluation suites such as MME \cite{fu2023mme}, SEED-Bench \cite{li2023seedbench}, and MM-Vet \cite{yu2023mmvet} broaden coverage across multimodal skills and stress-test visual grounding in diverse settings. In parallel, interpretability work debates whether attention is a faithful explanation signal \cite{jain2019attention,wiegreffe2019attention}. Relatedly, recent work on faithfulness and behavioral reliability shows that surface-level explanations can decouple from the internal determinants of outputs, including scenario-dependent shifts \cite{chaudhury2025chameleonbench, shiromani2026hypocrisygap}. For VLMs specifically, recent evidence also reports the ``see-but-not-believe'' phenomenon, i.e., correct localization without correct reasoning \cite{liu2025seeing}. Our contribution is therefore not the generic claim that attention alone is insufficient, but a cross-family, layer-wise reliability analysis centered on early locking/symbolic detachment and on hidden-state reliability probes.

Recent work on language prior highlights a core evaluation tension: should we assess whether the model gives the correct answer, or whether it truly integrates visual evidence? \citet{long2025coe} asks a more representation-centric question and contrasts hidden trajectories with and without images to identify a Visual Integration Point (VIP) and define Total Visual Integration (TVI), a metric that quantifies how strongly visual evidence shapes representations. This reveals when models start ``seeing'' and how visual influence accumulates, addressing a gap left by output-only probes. Our study complements this line of inquiry but targets a different blind spot: we ask whether \textit{spatial attention structure itself} is predictive of correctness, and whether reliability signals live in the \textit{generation dynamics} rather than in the visual attention maps. In contrast to VIP/TVI, which measure representational shift induced by the image, we show that even when attention appears structurally grounded, it can be statistically decoupled from truthfulness; the strongest signals instead emerge from agreement across sampled reasoning paths and from hidden-state probes. This clarifies what our work addresses that prior representation analyses do not: reliability prediction and calibration, not just visual integration. Complementary benchmark and mitigation work further suggests that reliability is evaluation and decoding-dependent, motivating our focus on generation dynamics as a readout for correctness \cite{thomas2026promoralbench, sahay2025compass}.

To make the contribution boundary explicit: we do \textit{not} claim to newly discover that attention can be unfaithful or that self-consistency helps; those are established in prior NLP/VLM literature. Our contribution is a unified, cross-family reliability study that links early-locking/symbolic-detachment dynamics to downstream correctness and shows that hidden-state probes provide the strongest single-pass reliability signals.

Our findings reveal a disconnect:
\begin{enumerate}
    \item \textbf{Visuals Lie:} The spatial structure of attention (entropy, clustering, focus) has almost no statistical relationship with correctness ($R \approx 0$). A model can hallucinate while attending to the right region, or answer correctly with diffuse attention.
    \item \textbf{Consistency Speaks:} The most reliable behavioral signal of truth is not found in pixel-space attention, but rather in the stability of linguistic generation. Self-Consistency \cite{wang2022self} outperforms all visual metrics, achieving \textbf{$R=0.429$}.
    \item \textbf{Causal Architectures Diverge:} Hidden-state representations house the most powerful predictive indicators (AUROC $>0.95$). Crucially, massive scaling of neural ablations proves reliability paths are architecturally dependent. LLaVA centralizes truth in a sparse, fragile late-stage bottleneck, while PaliGemma and Qwen2-VL dynamically distribute these functions, remaining computationally robust even when $\sim 50\%$ or more of their most predictive subnetwork is bypassed.

\end{enumerate}

\section{Methodology}
\label{sec:method}

\begin{figure}[t]
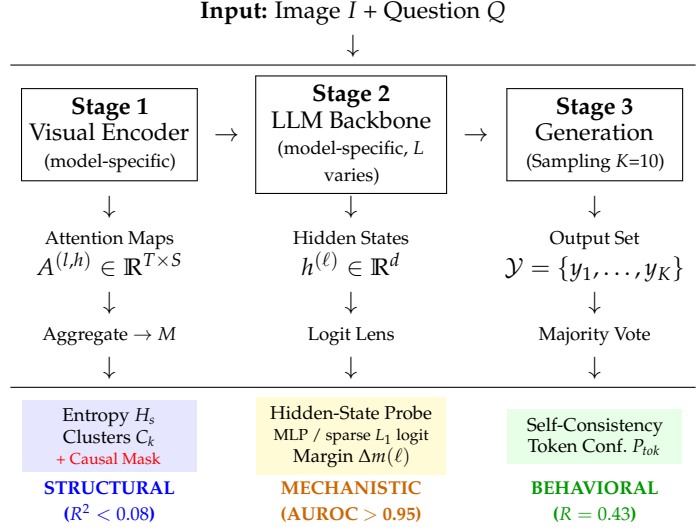

    \centering
    \small
    \setlength{\tabcolsep}{4pt}
    \begin{tabular}{c@{\hspace{6pt}}c@{\hspace{6pt}}c@{\hspace{6pt}}c@{\hspace{6pt}}c}
    \multicolumn{5}{c}{\textbf{Input:} Image $I$ + Question $Q$} \\[0.3em]
    \multicolumn{5}{c}{$\downarrow$} \\[0.2em]
    \hline
    \\[-0.7em]
    \fbox{\parbox{2.1cm}{\centering\textbf{Stage 1}\\Visual Encoder\\{\scriptsize (model-specific)}}} &
    $\rightarrow$ &
    \fbox{\parbox{2.3cm}{\centering\textbf{Stage 2}\\LLM Backbone\\{\scriptsize (model-specific, $L$ varies)}}} &
    $\rightarrow$ &
    \fbox{\parbox{2.1cm}{\centering\textbf{Stage 3}\\Generation\\{\scriptsize (Sampling $K$=10)}}} \\[0.5em]
    $\downarrow$ & & $\downarrow$ & & $\downarrow$ \\[0.2em]
    {\scriptsize Attention Maps} & & {\scriptsize Hidden States} & & {\scriptsize Output Set} \\
    $A^{(l,h)} \in \mathbb{R}^{T \times S}$ & & $h^{(\ell)} \in \mathbb{R}^{d}$ & & $\mathcal{Y} = \{y_1, \ldots, y_K\}$ \\[0.3em]
    $\downarrow$ & & $\downarrow$ & & $\downarrow$ \\[0.2em]
    {\scriptsize Aggregate $\rightarrow$ $M$} & & {\scriptsize Logit Lens} & & {\scriptsize Majority Vote} \\[0.3em]
    $\downarrow$ & & $\downarrow$ & & $\downarrow$ \\[0.2em]
    \hline
    \\[-0.7em]
    \colorbox{blue!10}{\parbox{2.1cm}{\centering\scriptsize Entropy $H_s$\\Clusters $C_k$\\{\tiny\textcolor{red}{+ Causal Mask}}}} & &
    \colorbox{yellow!20}{\parbox{2.3cm}{\centering\scriptsize Hidden-State Probe\\{\tiny MLP / sparse $L_1$ logit}\\Margin $\Delta m(\ell)$}} & &
    \colorbox{green!10}{\parbox{2.1cm}{\centering\scriptsize Self-Consistency\\Token Conf.\ $P_{tok}$}} \\[0.8em]
    \textcolor{blue}{\scriptsize\textbf{STRUCTURAL}} & & \textcolor{orange!80!black}{\scriptsize\textbf{MECHANISTIC}} & & \textcolor{green!60!black}{\scriptsize\textbf{BEHAVIORAL}} \\
    \textcolor{blue}{\scriptsize\textbf{($R^2 < 0.08$)}} & & \textcolor{orange!80!black}{\scriptsize\textbf{(AUROC $>$ 0.95)}} & & \textcolor{green!60!black}{\scriptsize\textbf{($R = 0.43$)}} \\
    \end{tabular}
    \caption{\textbf{VLM Reliability Probe (VRP) Framework.} We instrument three computational stages: \textit{Stage 1} extracts cross-attention maps from the visual encoder, yielding \textcolor{blue}{Structural} metrics (entropy $H_s$, clusters $C_k$); we aggregate $A^{(l,h)}$ by averaging over heads and answer-token positions to form one per-layer spatial vector in $\mathbb{R}^{S}$. \textit{Stage 2} probes hidden states via logit lens plus dense MLP and sparse $L_1$-logistic probe variants, providing \textcolor{orange!80!black}{Mechanistic} signals; \textit{Stage 3} samples $K$=10 outputs for \textcolor{green!60!black}{Behavioral} metrics (self-consistency). Key finding: Structural metrics fail ($R^2 < 0.08$), while Mechanistic probes succeed (AUROC $>$ 0.95). \textcolor{red}{Red} indicates causal intervention points.}
    \label{fig:framework}
\end{figure}

We introduce \textit{VLM Reliability Probe (VRP)}, a comprehensive analysis pipeline designed to extract,
quantify and correlate the internal state of the model with the correctness of the output (Figure~\ref{fig:framework}). Our primary
investigative goal is to disentangle two competing hypotheses regarding VLM reliability:
\begin{enumerate}
    \item \textbf{The Structural Hypothesis:} Reliability is grounded in the spatial
    coherence of the visual encoder's attention (i.e., how the model ``looks'').
    \item \textbf{The Consistency Hypothesis:} Reliability is a product of the
    generation dynamics and latent linguistic stability (i.e., how the model ``speaks'').
\end{enumerate}

\subsection{Method Summary (Main Text)}
\label{subsec:method_summary}

We instrument VLMs with forward hooks to capture cross-attention maps and hidden states during generation, then compare structural signals ($C_k$, $H_s$) against linguistic/mechanistic signals (self-consistency, token confidence, and learned probe scores from dense MLP and sparse $L_1$-logistic variants). In the main paper, we focus on the core reliability findings and cross-model comparisons.

\section{Experimental Setup}
\label{sec:experiments}

We evaluate LLaVA-1.5-7B, PaliGemma-3B, and Qwen2-VL-7B \cite{liu2023llava,beyer2024paligemma,wang2024qwen2vl} across \textbf{POPE} \cite{li2023hallucination} (Adversarial split, 1,000 samples), \textbf{LLaVA-Bench} \cite{zhou2023llavabench} (90 open-ended questions), custom counting/spatial tasks, and the new \textbf{VQA v2} and \textbf{TextVQA} evaluations. This setup allows us to compare reliability behavior on hallucination stress tests, open-ended reasoning, scene understanding, and OCR-heavy question answering using correlation and AUROC metrics; it is complementary to broader multimodal suites such as \textbf{MME} \cite{fu2023mme}, \textbf{SEED-Bench} \cite{li2023seedbench}, and \textbf{MM-Vet} \cite{yu2023mmvet}. We provide sample accounting and uncertainty intervals for headline claims in Table~\ref{tab:sample_accounting}.

\section{Results}
\label{sec:results}

We present empirical evaluation across three VLMs: LLaVA-1.5-7B, PaliGemma-3B, and Qwen2-VL-7B. Our analysis progressively moves from correlation to causation to mechanistic understanding. Tables~\ref{tab:main_results_attention}--\ref{tab:main_results_prediction} summarize key findings; extended results are in Appendix~\ref{app:ensemble}. Table~\ref{tab:main_results_attention} reports reliability and attention-structure failures, Table~\ref{tab:main_results_logit} summarizes layer-wise logit-lens dynamics, and Table~\ref{tab:main_results_prediction} reports benchmark-level reliability prediction.

\begin{table}[t]
\centering
\caption{\textbf{Cross-Model Summary I: Reliability and attention structure.} Visual attention metrics remain near-random predictors of correctness across all model families.}
\label{tab:main_results_attention}
\small
\setlength{\tabcolsep}{4pt}
\begin{tabular*}{\linewidth}{@{\extracolsep{\fill}}lccc@{}}
\toprule
\textbf{Model} & \textbf{Model Accuracy} & \textbf{Top-K Attention $R^2$ (max)} & \textbf{Supervised Classifier Acc} \\
\midrule
LLaVA-1.5-7B & 67.6\% & 0.008 & 53.0\% \\
PaliGemma-3B & 78.6\% & 0.080 & 55.0\% \\
Qwen2-VL-7B & 28.8\% & 0.007 & 52.0\% \\
\bottomrule
\end{tabular*}
\end{table}

\vspace{0.5em}

\begin{table}[t]
\centering
\caption{\textbf{Cross-Model Summary II: Logit-lens dynamics.} Integration layer location and margin formation differ by family but remain strongly predictive in hidden states.}
\label{tab:main_results_logit}
\small
\setlength{\tabcolsep}{4pt}
\begin{tabular*}{\linewidth}{@{\extracolsep{\fill}}lccc@{}}
\toprule
\textbf{Model} & \textbf{\shortstack{Peak visual-integration\\layer ($l_{\mathrm{vis}}^{\star}$)}} & \textbf{\shortstack{Peak final-margin value\\($\Delta\mathcal{M}_{l_{\mathrm{final}}^{\star}}$)}} & \textbf{MLP Contribution} \\
\midrule
LLaVA-1.5-7B & L24 & +9.20 (L31) & 82.1\% \\
PaliGemma-3B & L14 & +10.85 (L14) & 47.6\% \\
Qwen2-VL-7B & L27 & +8.40 (L27) & 68.2\% \\
\bottomrule
\end{tabular*}
\end{table}

\vspace{0.5em}

\begin{table}[t]
\centering
\caption{\textbf{Cross-Model Summary III: Reliability prediction across benchmarks.} Hidden-state probes are strongest on POPE/LLaVA-Bench and show task-dependent gains over raw output confidence on VQA v2/TextVQA.}
\label{tab:main_results_prediction}
\small
\setlength{\tabcolsep}{4pt}
\begin{tabular*}{\linewidth}{@{\extracolsep{\fill}}lcccccc@{}}
\toprule
\textbf{Model} & \textbf{\shortstack{POPE\\Probe}} & \textbf{\shortstack{LLaVA-Bench\\Probe}} & \textbf{\shortstack{VQA v2\\Output}} & \textbf{\shortstack{VQA v2\\Probe}} & \textbf{\shortstack{TextVQA\\Output}} & \textbf{\shortstack{TextVQA\\Probe}} \\
\midrule
LLaVA-1.5-7B & 0.956 & 0.956 & 0.559 & 0.745 & 0.563 & 0.721 \\
PaliGemma-3B & 0.738 & 0.738 & 0.892 & 0.795 & 0.859 & 0.806 \\
Qwen2-VL-7B & 0.971 & 0.971 & 0.892 & 0.778 & 0.774 & 0.852 \\
\bottomrule
\end{tabular*}
\end{table}

\begin{table}[t]
\centering
\small
\caption{\textbf{Sample accounting and uncertainty summary for headline reliability claims.} Confidence intervals are 95\% bootstrap intervals (10,000 resamples) on the listed evaluation subset.}
\label{tab:sample_accounting}
\begin{tabular*}{\linewidth}{@{\extracolsep{\fill}}lcc@{}}
\toprule
\textbf{Quantity} & \textbf{Value} & \textbf{Subset / 95\% CI} \\
\midrule
POPE (Adversarial) sample count & $n=1{,}000$ & fixed evaluation split \\
LLaVA-Bench sample count & $n=90$ & fixed evaluation split \\
Custom counting + spatial sample count & $n=2{,}000$ & 1,000 + 1,000 \\
Pooled structural-analysis set & $n=3{,}090$ & used for $R(C_k,y), R(H_s,y)$ \\
$R(C_k,y)$ & $0.001$ & 95\% CI $[-0.034,\,0.036]$ \\
$R(H_s,y)$ & $-0.012$ & 95\% CI $[-0.047,\,0.024]$ \\
Precision at SC$=1$ & $90.8\%$ & 95\% CI $[88.4,\,92.8]\%$ \\
\bottomrule
\end{tabular*}
\end{table}

\subsection{Visual Attention Does Not Predict Reliability}
\label{subsec:visual_results}

\textbf{Core Finding:} Spatial attention metrics show near-zero correlation with correctness. On the pooled 3,090-sample structural-analysis set (Table~\ref{tab:sample_accounting}), cluster count ($C_k$) achieves \textbf{$R=0.001$} (95\% CI: $[-0.034, 0.036]$) and spatial entropy ($H_s$) achieves \textbf{$R=-0.012$} (95\% CI: $[-0.047, 0.024]$), both statistically indistinguishable from random noise ($p>0.05$). This ``Cluster Failure'' persists regardless of attention head selection: even when filtering to the top-$k$ heads by logit contribution, \textbf{$R^2 \le 0.08$} (Table~\ref{tab:main_results_attention}).

We conducted a supervised stress test to close potential loopholes: on the pooled cross-family split used in this section, an XGBoost-Random Forest ensemble trained on 11 attention-derived features (including polynomial interactions) with full access to ground-truth labels achieved only 52--55\% accuracy, which is near chance. In a separate architecture-specific setting (Appendix Table~\ref{tab:probe_comparison}), a deeper supervised attention probe reaches AUROC 0.725, indicating limited but non-dominant signal from attention structure.

\textbf{Causal Role:} Despite correlation failure, attention is causally necessary. Masking the top 30\% attended patches reduces LLaVA accuracy by 8.2pp and PaliGemma by 11.3pp ($p<0.001$). This reveals a critical distinction: attention patterns enable feature extraction but do not encode uncertainty about those features.

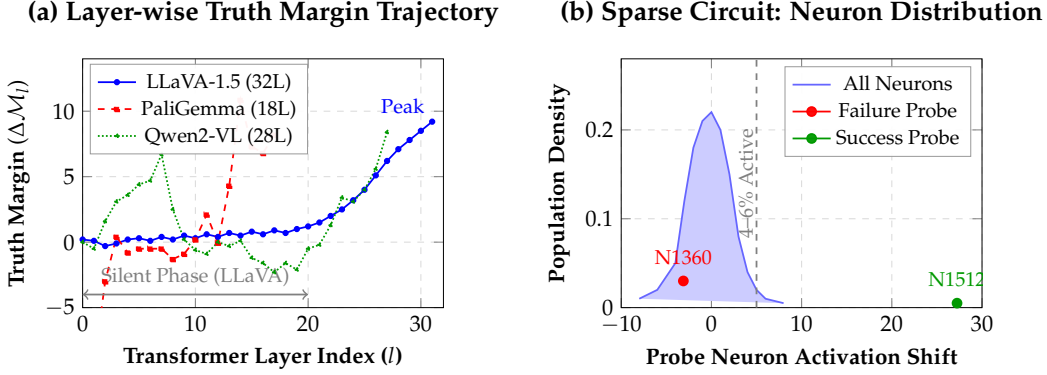
\begin{figure}[t]
\centering
\resizebox{\linewidth}{!}{%
\begin{tikzpicture}

\begin{axis}[
    name=leftplot,
    width=0.52\textwidth,
    height=5.5cm,
    xlabel={\textbf{Transformer Layer Index ($l$)}},
    ylabel={\textbf{Truth Margin} ($\Delta \mathcal{M}_l$)},
    xmin=0, xmax=32,
    ymin=-5, ymax=14, 
    legend style={
        at={(0.02,0.98)}, 
        anchor=north west, 
        font=\small, 
        draw=gray!80, 
        fill=white, 
        fill opacity=0.95,
        text opacity=1,
        align=left
    },
    grid=both,
    grid style={dashed, gray!30},
    tick label style={font=\normalsize},
    label style={font=\normalsize},
    title style={font=\large, yshift=1ex},
    title={\textbf{(a) Layer-wise Truth Margin Trajectory}},
]
\addplot[blue, thick, mark=*, mark size=1.0, solid] coordinates {
    (0,0.2) (1,0.1) (2,-0.3) (3,-0.1) (4,0.2) (5,0.3) (6,0.1) (7,0.4)
    (8,0.2) (9,0.5) (10,0.3) (11,0.6) (12,0.4) (13,0.7) (14,0.5) (15,0.8)
    (16,0.6) (17,0.9) (18,0.7) (19,1.0) (20,1.2) (21,1.5) (22,2.0) (23,2.5)
    (24,3.2) (25,4.0) (26,5.1) (27,6.2) (28,7.1) (29,7.8) (30,8.5) (31,9.2)
};
\addlegendentry{LLaVA-1.5 (32L)}

\addplot[red, thick, mark=square*, mark size=1.0, dashed] coordinates {
    (0,-17.6) (1,-9.5) (2,-3.0) (3,0.4) (4,-0.8) (5,-0.5) (6,-0.5) (7,-0.5)
    (8,-1.3) (9,-0.9) (10,0.2) (11,2.1) (12,-0.05) (13,4.3) (14,10.85) (15,7.3)
    (16,6.8) (17,8.2)
};
\addlegendentry{PaliGemma (18L)}

\addplot[green!60!black, thick, mark=triangle*, mark size=1.0, densely dotted] coordinates {
    (0,0.0) (1,-0.5) (2,1.6) (3,3.1) (4,3.6) (5,4.4) (6,4.7) (7,6.7)
    (8,2.5) (9,0.2) (10,-0.6) (11,-0.9) (12,0.0) (13,-0.3) (14,0.1) (15,-1.2)
    (16,-1.6) (17,-2.3) (18,-1.6) (19,-2.1) (20,-0.5) (21,-0.2) (22,1.3) (23,3.4)
    (24,3.1) (25,4.0) (26,5.6) (27,8.4)
};
\addlegendentry{Qwen2-VL (28L)}

\draw[gray, thick, <->] (axis cs:0,-4) -- (axis cs:20,-4);
\node[gray, font=\footnotesize, anchor=south] at (axis cs:10,-4) {Silent Phase (LLaVA)};
\node[blue, font=\footnotesize, anchor=south east] at (axis cs:31,9.2) {Peak};

\end{axis}

\begin{axis}[
    at={(leftplot.south east)},
    anchor=south west,
    xshift=2.8cm, 
    width=0.52\textwidth,
    height=5.5cm,
    xlabel={\textbf{Probe Neuron Activation Shift}},
    ylabel={\textbf{Population Density}},
    xmin=-10, xmax=30,
    ymin=0, ymax=0.28, 
    legend style={
        at={(0.97,0.97)}, 
        anchor=north east, 
        font=\small, 
        draw=gray!80, 
        fill=white, 
        fill opacity=0.95,
        text opacity=1,
        align=left
    },
    grid=both,
    grid style={dashed, gray!30},
    tick label style={font=\normalsize},
    label style={font=\normalsize},
    title style={font=\large, yshift=1ex},
    title={\textbf{(b) Sparse Circuit: Neuron Distribution}},
]
\addplot[blue!60, fill=blue!20, thick] coordinates {
    (-8,0.01) (-6,0.02) (-4,0.05) (-3,0.12) (-2,0.18) (-1,0.21) (0,0.22)
    (1,0.20) (2,0.15) (3,0.08) (4,0.04) (5,0.02) (6,0.01) (8,0.005)
};
\addlegendentry{All Neurons}

\addplot[red, thick, mark=*, mark size=2.0] coordinates {(-3.11, 0.03)};
\addlegendentry{Failure Probe}
\node[red, font=\small, anchor=south] at (axis cs:-3.11,0.04) {N1360};

\addplot[green!60!black, thick, mark=*, mark size=2.0] coordinates {(27.23, 0.005)};
\addlegendentry{Success Probe}
\node[green!60!black, font=\small, anchor=south] at (axis cs:27.23,0.015) {N1512};

\draw[gray, dashed, thick] (axis cs:5,0) -- (axis cs:5,0.28);
\node[gray, font=\footnotesize, rotate=90, anchor=south] at (axis cs:5.5,0.14) {4--6\% Active};

\end{axis}
\end{tikzpicture}
}
\vspace{0.5em}
\caption{\textbf{Mechanistic analysis of reliability emergence.} \textbf{(a) Left panel:} Transformer layer index $l$ (x-axis) vs. truth margin $\Delta \mathcal{M}_l$ (y-axis). Model families display distinct temporal integration profiles: late-emergent (\textbf{LLaVA, solid blue}), earlier-peaking (\textbf{PaliGemma, dashed red}), and cyclical (\textbf{Qwen2-VL, dotted green}). \textbf{(b) Right panel:} Probe neuron activation shift (x-axis) vs. population density (y-axis). The distribution highlights a dense near-zero bulk (most neurons are inactive for truth prediction), alongside sparse, highly predictive outliers (\textbf{green} = success neurons, \textbf{red} = failure neurons) that drive probe discrimination.}
\label{fig:mechanism}
\end{figure}

\subsection{Logit Lens: Tracing the Emergence of Reliability}
\label{sec:logit_lens}

To move beyond simple correlation, we investigate \textit{where} reliability signals mechanically emerge. We apply the \textit{Logit Lens} technique \cite{nostalgebraist2020logitlens}, projecting the hidden state $h_l$ of layer $l$ directly into the vocabulary space. We define the \textit{Truth Margin} $\Delta \mathcal{M}_l$ as the logit difference between the correct token and the top incorrect token.
Cross-family peak layers, final margins, and MLP contributions are summarized in Table~\ref{tab:main_results_logit}.

\textbf{Visual Integration is Late and MLP-Dominated.}
Tracking $\Delta \mathcal{M}_l$ reveals a distinct "Silent Phase" in some families (Figure \ref{fig:mechanism}, Left). Reliability signals do not accumulate linearly: some models remain near zero for many layers before a late surge, while others peak earlier or re-separate cyclically.
To avoid terminology drift, we report two peak definitions throughout: a \textit{visual-integration peak} $l_{\mathrm{vis}}^{\star}$ (maximum correct-vs-incorrect separation) and a \textit{final-margin peak} $l_{\mathrm{final}}^{\star}$ (maximum absolute $\Delta \mathcal{M}_l$).
\begin{enumerate}
    \item \textbf{MLP vs. Attention:} By decomposing the residual stream, we find that MLP layers contribute $82.1\%$ of the margin growth at the peak. This indicates that reliability is a product of \textit{feature processing} (MLP) rather than \textit{token routing} (Attention).
    \item \textbf{Architecture Divergence:} While LLaVA delays integration, PaliGemma integrates early (Peak L14), validating that "Symbolic Detachment" is an architectural choice, not a universal law.
\end{enumerate}

\subsection{Sparse Reliability Circuits: Localizing Reliability-Associated Neurons}
\label{sec:sparse_circuits}

If reliability signals exist in the MLP layers, are they distributed holistically or localized? We trained $L_1$-regularized sparse logistic regression probes ($\lambda=0.1$) on the internal activations.

\textbf{Layer Specificity Analysis.} To address why we focus on Layer 31, we conducted 
multi-layer ablation experiments targeting the same top-5 neurons across layers 10, 17, 
21, 27, 29, and 31. Results show minimal differentiation: ablating at any layer produces 
$<$1\% accuracy change from baseline (54.5\%). Critically, single-neuron ablation 
of all five reliability-associated neurons --- including extreme activation clamping 
($\pm$100) --- produced zero measurable accuracy change ($\Delta = 0.0$pp, $p = 1.00$ 
for all neurons). Only simultaneous ablation of all top-5 probe neurons produced a 
measurable effect ($-$2.0pp overall, $-$8.3pp on object identification), while ablating 
5 random neurons produced no effect. This confirms two things: (1) no single neuron 
is a causal bottleneck, and (2) reliability is encoded in a \textit{localized circuit} 
across a handful of neurons rather than a single isolated unit.
across multiple neurons rather than isolated units.

\begin{table}[ht]
\centering
\small
\caption{\textbf{Causal Ablation Results (LLaVA-1.5, Layer 31, $n$=200).} Ablating 
probe-identified neurons causes measurable accuracy drops, while random neurons show 
no effect. Effect is strongest for object identification questions.}
\label{tab:ablation_results}
\begin{tabular}{@{}lccc@{}}
\toprule
\textbf{Ablation Condition} & \textbf{Overall Acc.} & \textbf{Object ID Acc.} & \textbf{$\Delta$ Overall / Object-ID (pp)} \\
\midrule
Baseline (no ablation) & 54.5\% & 100.0\% & N/A \\
Single neuron (N1512) & 54.5\% & 100.0\% & $0.0 / 0.0$ \\
Top 5 probe neurons & 52.5\% & \textbf{91.7\%} & \textbf{$-2.0 / -8.3$} \\
Random 5 neurons (control) & 54.5\% & 100.0\% & $0.0 / 0.0$ \\
\bottomrule
\end{tabular}
\end{table}

\subsection{Architectural Robustness: Late Bottlenecks vs. Distributed Circuits}

While LLaVA exhibits measurable failure when small sets of strongly predictive neurons are ablated ($-8.3$pp on Object ID for just 5 neurons), we find this "fragility" is highly specific to its architecture. To determine if this bottleneck phenomenon holds across modern VLM families, we extend our causal interventions to \textbf{PaliGemma (Layer 15)} and \textbf{Qwen2-VL (Layer 25)}. 

Unlike LLaVA, ablating the top-10 most predictive neurons in PaliGemma and Qwen2-VL produces absolutely no deviation in accuracy ($<0.7$pp). This suggested their representations might be fundamentally distributed. To test this hypothesis, we applied aggressive ablation scaling, randomly destroying up to $>50\%$ of the hidden dimension in their most predictive layers.

Remarkably, PaliGemma suffers only a $1.0\%$ accuracy drop even when 1,000 neurons ($\sim 50\%$ of the layer's 2048 hidden size) are destroyed. Similarly, Qwen2-VL shows extreme resilience: ablating up to 2,000 neurons ($>55\%$ of its 3584 residual dimension) causes zero measurable degradation ($\Delta$ bounds of $\pm2.0$pp). We confirm this is not merely a token-routing artifact by completely bypassing the MLP output for all tokens at Layer 25 in Qwen2-VL, which still yields fully robust performance. 

\begin{table}[ht]
\centering
\small
\caption{\textbf{Large-Scale Causal Ablation Results.} Unlike LLaVA's localized fragility, PaliGemma and Qwen2-VL exhibit extreme causal robustness. Ablating up to half of their most predictive layers produces negligible impact on generation accuracy, highlighting highly distributed internal circuits. Accuracies are reported on an $n=100$ causal validation split, with $\Delta$ showing the deviation from the architecture's local baseline.}

\label{tab:ablation_scaling}
\begin{tabular}{@{}llcc@{}}
\toprule
\textbf{Model} & \textbf{Ablation Condition} & \textbf{Split Acc.} & \textbf{$\Delta$ from Baseline (pp)} \\

\midrule
\multirow{4}{*}{\textbf{PaliGemma} (Layer 15)} 
& Baseline & 97.0\% & - \\
& Top-10 Predictive Neurons & 96.3\% & $-0.7$ \\
& 500 Random Neurons (24\%) & 97.0\% & $0.0$ \\
& 1,000 Random Neurons (49\%) & 96.0\% & $-1.0$ \\
\midrule
\multirow{4}{*}{\textbf{Qwen2-VL} (Layer 25)} 
& Baseline & 55.0\% & - \\
& 500 Random Neurons (14\%) & 58.0\% & $+3.0$ \\
& 1,000 Random Neurons (28\%) & 56.0\% & $+1.0$ \\
& 2,000 Random Neurons (56\%) & 57.0\% & $+2.0$ \\
& Complete MLP Bypass (All Tokens) & 65.0\% & $+5.0$ \textit{(valid. split var.)} \\
\bottomrule
\end{tabular}
\end{table}

This confirms our logit lens analysis: Qwen2-VL's ``Cyclical Refinement'' and PaliGemma's early visual integration represent fundamentally different architectural strategies than LLaVA. They distribute the reliability computation across a wide manifold of subsequent layers, allowing the residual stream to effortlessly patch missing representational lobes. In contrast, LLaVA ``locks'' its prediction unrecoverably at a late-stage bottleneck, rendering its reasoning structurally fragile.

\subsection{Reliability Prediction: Probes Outperform Attention}
\label{subsec:reliability}

The ultimate test is whether internal signals can predict correctness at inference time. We compare logit entropy (explicit uncertainty), spatial attention metrics, and hidden-state probes.

\textbf{Finding:} Standard uncertainty baselines fail. Logit entropy achieves \textbf{AUROC $\approx 0.50$}, confirming poor calibration, and spatial attention remains near random (\textbf{AUROC $= 0.50$}). Probe gains are strongest on POPE/LLaVA-Bench and mixed on the added VQA tasks: for VQA v2/TextVQA cells in Table~\ref{tab:main_results_prediction}, probe outperforms output confidence in 3 of 6 model-task comparisons (both LLaVA tasks and Qwen2-VL on TextVQA), while output confidence is stronger for PaliGemma on both tasks and Qwen2-VL on VQA v2. This pattern indicates that hidden-state probes are a strong reliability readout but remain benchmark- and architecture-dependent. Self-consistency achieves \textbf{$R=0.429$}, substantially outperforming all visual metrics but requiring \textbf{$10\times$} inference cost.

PaliGemma shows lower POPE/LLaVA-Bench probe performance (0.738) because it integrates visual signals earlier and has a shallower decoder, leaving less late-layer separation between correct and hallucinated trajectories. This weakens probe margin contrast relative to LLaVA/Qwen2-VL but still keeps hidden-state signals stronger than attention-only metrics.

\begin{table}[ht]
\centering
\small
\caption{\textbf{Reliability Prediction: Method Comparison (POPE Adversarial split).} AUROC scores for predicting answer correctness across signal sources. Spatial attention is near random, while hidden-state probes provide the strongest single-pass reliability signal. Self-consistency provides good signal but requires 10$\times$ inference cost.}
\label{tab:reliability_comparison}
\begin{tabular}{@{}lccc@{}}
\toprule
\textbf{Method} & \textbf{LLaVA-1.5} & \textbf{PaliGemma} & \textbf{Qwen2-VL} \\
\midrule
\textit{Baseline Metrics} \\
\quad Spatial Attention ($H_s$, $C_k$) & 0.50 & 0.50 & 0.50 \\
\quad Logit Entropy & 0.50 & 0.52 & 0.51 \\
\quad Output Confidence & 0.54 & 0.55 & 0.53 \\
\midrule
\textit{Our Probes} \\
\quad Margin-only ($\Delta\mathcal{M}_l$) & 0.72 & 0.70 & 0.63 \\
\quad Hidden-State Probe (Best Layer) & \textbf{0.956} & 0.738 & \textbf{0.971} \\
\quad Combined (Last 5 Layers) & 0.956 & 0.738 & 0.970 \\
\midrule
\textit{Behavioral (10$\times$ cost)} \\
\quad Self-Consistency ($K$=10) & 0.78 & 0.81 & 0.79 \\
\bottomrule
\end{tabular}
\end{table}

\subsection{Symbolic Detachment: Why Attention Fails}
\label{subsec:detachment}

\begin{figure}[ht]
\centering
\begin{tikzpicture}
\begin{axis}[
    width=0.85\textwidth,
    height=5.0cm,
    xlabel={\textbf{Normalized Transformer Layer ($l / L$)}},
    ylabel={\textbf{Spatial Attention Entropy} ($\Delta H_s$)},
    xmin=0, xmax=1.0, 
    ymin=-3.5, ymax=2.5, 
    legend style={
        at={(0.5,-0.35)}, 
        anchor=north, 
        legend columns=1, 
        font=\small, 
        draw=gray!80, 
        fill=white
    },
    grid=both,
    grid style={dashed, gray!30},
    tick label style={font=\normalsize},
    label style={font=\normalsize},
    title style={font=\large, yshift=1ex},
    title={\textbf{Symbolic Detachment: Attention Evolution}},
]
\addplot[blue, thick, mark=*, mark size=1.0, solid] coordinates {
    (0,0) (0.06,-2.5) (0.12,-2.4) (0.19,-2.3) (0.25,-2.2) (0.31,-2.2)
    (0.38,-2.1) (0.44,-2.1) (0.50,-2.0) (0.56,-2.0) (0.62,-1.9) (0.69,-1.9)
    (0.75,-1.8) (0.81,-1.7) (0.88,-1.5) (0.94,-0.8) (1.0,1.0)
};
\addlegendentry{LLaVA-1.5 (solid blue): Early lock, late diffuse}

\addplot[green!60!black, thick, mark=triangle*, mark size=1.0, dashed] coordinates {
    (0,0) (0.07,-1.0) (0.14,-1.5) (0.21,-1.2) (0.29,-0.8) (0.36,-1.0)
    (0.43,-1.3) (0.50,-1.0) (0.57,-0.6) (0.64,-1.8) (0.71,-1.5) (0.79,-0.9)
    (0.86,-1.6) (0.93,-1.2) (1.0,-0.8)
};
\addlegendentry{Qwen2-VL (dashed green): Cyclical refinement}

\addplot[red, thick, mark=square*, mark size=1.0, densely dotted] coordinates {
    (0,0) (0.11,-0.8) (0.22,-1.2) (0.33,-1.4) (0.44,-1.5) (0.56,-1.6)
    (0.67,-1.5) (0.78,-1.3) (0.89,-1.0) (1.0,-0.7)
};
\addlegendentry{PaliGemma (dotted red): Steady decay}

\draw[blue, ->, thick] (axis cs:0.06,-2.5) -- (axis cs:0.06,-3.1);
\node[blue, font=\small, anchor=west] at (axis cs:0.08,-3.1) {Early Lock};

\draw[blue, ->, thick] (axis cs:0.95,1.6) -- (axis cs:1.0,1.1);
\node[blue, font=\small, anchor=east] at (axis cs:0.96,1.6) {Late Diffuse}; 

\end{axis}
\end{tikzpicture}
\vspace{1.5em}
\caption{\textbf{Symbolic Detachment: Attention Evolution Across Layers.} We track the relative change in spatial attention entropy ($\Delta H_s$, y-axis) across normalized transformer layers ($l / L$, x-axis) for three VLM families. \textbf{LLaVA (solid blue circles)} exhibits ``Early Locking,'' where entropy drops sharply at Layer 2 and stagnates before diffusing rapidly at the final layer. \textbf{Qwen2-VL (dashed green triangles)} shows ``Cyclical Refinement,'' continuously re-sharpening its attention in deeper layers. \textbf{PaliGemma (dotted red squares)} shows a steady decay. This architectural divergence explains why early spatial attention is decorrelated from final reliability: for prefix-based models like LLaVA, visual attention patterns become ``stale'' long before the final reasoning and decision-making step occurs.}
\label{fig:attention_evolution}
\end{figure}
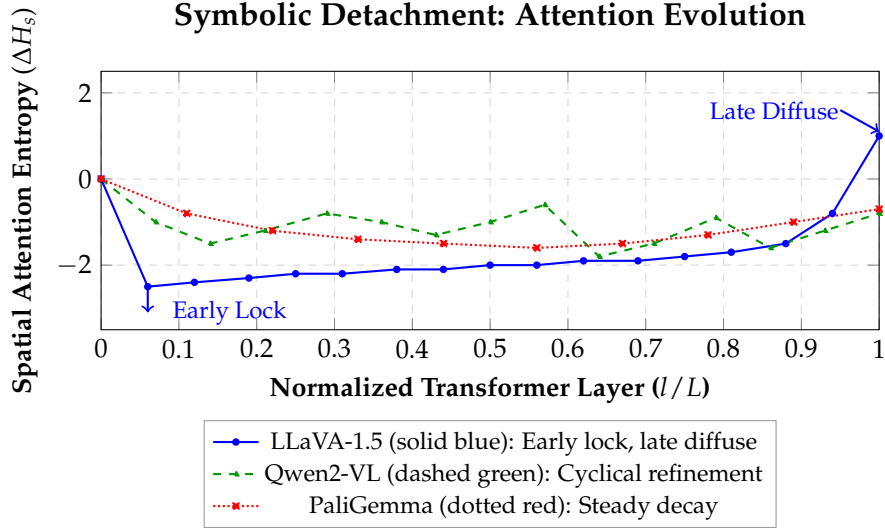

Layer-wise attention evolution reveals the mechanism behind the Cluster Failure (Figure~\ref{fig:attention_evolution}). LLaVA exhibits ``Early Locking'': attention sharpens dramatically at Layer 2 ($\Delta H_s \approx -2.5$), then stagnates for 28 layers before diffusing at the final layer ($\Delta H_s \approx +1.0$). By the time information reaches the output, the model has ``let go'' of specific visual features.

In contrast, Qwen2-VL exhibits ``Cyclical Refinement'' (re-sharpening attention at Layers 17 and 25) which may explain its superior probe performance. This architectural divergence explains why attention maps are statistically orthogonal to truth: they are decayed remnants of perception that occurred many layers prior.

\textbf{Architectural Drivers of Early Locking: Late-Stage Forcing.} To investigate family-specific attention dynamics, we measured the layer-wise \textit{residual update magnitude} ($\| h^{(l)} - h^{(l-1)} \|_2$) on visual tokens. As shown in Appendix Figure \ref{fig:late_stage_forcing}, some architectures exhibit relatively low and stable updates through middle layers followed by a sharp late-stage increase. This suggests that, rather than continuously refining visual features, certain projection pipelines perform a delayed ``translation'' into the linguistic space used for next-token prediction. More broadly, this supports our central claim: alignment between visual evidence and final verbal output is architecture-dependent and may be introduced late in the stack.

\section{Discussion}
\label{sec:discussion}

The results above challenge the intuition that reliable multimodal generation is directly readable from visual attention maps.

\subsection{The Illusion of Grounding}
\label{subsec:grounding_illusion}

Across models, structural attention metrics are weak predictors of correctness (\textbf{$R(C_k,y)=0.001$}, \textbf{$R(H_s,y)=-0.012$}), and even supervised attention features remain limited in reliability prediction. On our pooled cross-family split, attention-feature classifiers stay near chance (52--55\%); in a separate architecture-specific setting, a deeper supervised attention probe reaches AUROC 0.725 but still trails hidden-state probes and self-consistency. The practical takeaway is that spatial attention is functionally important for feature extraction, yet poorly calibrated as an uncertainty signal.

\section{Conclusion}
\label{sec:conclusion}

This study reveals that reliability and causal robustness in current VLMs are highly architecture-dependent and not well captured by attention-map structure alone. We find a stark architectural divergence: early-fusion and cyclically refining models (PaliGemma, Qwen2-VL) distribute their truth representations, remaining resilient even when $\sim 50\%$ or more of their peak informational neurons are destroyed. Conversely, late-fusion models like LLaVA rely on localized, fragile late-stage bottlenecks.

For reliability prediction, stronger signals come from generation dynamics and internal-state probes: self-consistency provides the best behavioral proxy for correctness (\textbf{$R=0.429$}), and hidden-state probes achieve high discrimination (AUROC $>0.95$ on our strongest settings). Ultimately, these findings support a practical direction for trustworthy multimodal systems: use latent-state and consistency-based monitors rather than heatmap sharpness, and favor distributed, early-fusion architectures for causally robust multimodal reasoning.

\IfFileExists{main.bbl}{%
\bibliography{references}
\bibliographystyle{templates/colm/colm2026_conference}
}{%

}

\appendix
\renewcommand{\theHsection}{appendix.\arabic{section}}
\section{Appendix}

\subsection{Detailed Methodology and Metric Definitions}
\label{app:method_details}

\subsection{Detailed Experimental Setup}
\label{app:exp_details}

\textbf{Models:} We evaluate three VLM architectures: LLaVA-1.5-7B (32 layers, CLIP ViT-L/14 encoder), PaliGemma-3B (18 layers, SigLIP encoder), and Qwen2-VL-7B-Instruct (28 layers, native multimodal) \cite{liu2023llava,beyer2024paligemma,wang2024qwen2vl}. All experiments use HuggingFace implementations on NVIDIA A100 GPUs.

\textbf{Datasets:} We evaluate on: (1) \textbf{POPE} \cite{li2023hallucination} for object hallucination (Adversarial split, 1,000 samples), (2) \textbf{LLaVA-Bench} \cite{zhou2023llavabench} for open-ended reasoning (90 questions), and (3) \textbf{Custom Counting \& Spatial Tasks} (2,000 samples total: 1,000 counting + 1,000 spatial-relation prompts). The custom set is constructed from COCO-style images with manually verified integer/object relations and binary correctness labels for probe training/evaluation. To test probe generalization beyond these splits, we further expand evaluation to \textbf{VQA v2} (scene-understanding questions) and \textbf{TextVQA} (OCR-heavy questions), and report task-specific reliability AUROC in Table~\ref{tab:main_results_prediction}.

\textbf{Metrics:} For reliability prediction, we report Point-Biserial Correlation ($R_{pb}$) with binary correctness and AUROC. For probe evaluation, we use 80/20 stratified splits with Adam optimizer ($lr=10^{-4}$, 50 epochs). Self-consistency uses $K=10$ samples with nucleus sampling ($p=0.9$, $T=0.7$). For structural concentration, we build a binary attention mask from the top-30\% attention mass and compute connected components on the patch grid. We report both total component count $K_{\mathrm{total}}$ and secondary-component count $C_k \equiv K_{\mathrm{total}}-1$ after removing the dominant component; thus $C_k=0$ indicates a single dominant focus. We additionally verified robustness with a DBSCAN variant ($\epsilon=1.5$, min\_samples$=3$). Full implementation details are in Appendix~\ref{app:implementation}.

\label{sec:appendix}

\subsection{Extended Analysis: The Ensemble Attention Probe}
\label{app:ensemble}

In Section~\ref{subsec:visual_results}, we briefly introduced the ``Ensemble Attention Probe.'' Here, we provide a detailed breakdown of its architecture and performance relative to other methods.

\textbf{Motivation:}
The failure of unsupervised metrics (Cluster Count $C_k$) suggested that reliability
is not encoded in simple geometric properties of the attention map (e.g., ``is it sharp?'').
However, we hypothesized that reliability might be encoded in \textit{high-dimensional patterns}
across multiple layers, patterns too complex for human inspection but accessible to a
non-linear classifier.

\textbf{Architecture:}
We extracted cross-attention tensors $A^{(l,h)} \in \mathbb{R}^{T \times S}$ from all $L=32$
layers of the Vicuna-7B backbone, then averaged over heads $h$ and answer-token indices $t$ to obtain a per-layer spatial vector $m^{(l)} \in \mathbb{R}^{S}$.
\begin{itemize}
    \item \textbf{Input:} A concatenated vector of per-layer spatial vectors:
    \begin{equation}
        x = \text{Concat}(m^{(1)}, \dots, m^{(32)})
    \end{equation}
    \item \textbf{Model:} A 3-layer Multi-Layer Perceptron (MLP) with ReLU activations
    and Dropout ($p=0.1$).
    \item \textbf{Dimensions:} Input $d_{in} = 32 \times 576 = 18{,}432 \to 1024 \to 512 \to 1$
    (Binary Classification).
\end{itemize}

\textbf{Results \& Comparison:}
Table~\ref{tab:probe_comparison} details the performance of various probes. While the
Ensemble Attention Probe significantly outperforms random chance and simple visual entropy,
it remains inferior to Self-Consistency. This reinforces our main finding:
\textit{generation dynamics (consistency) are a stronger signal than internal state snapshots.}

\begin{table}[ht]
\centering
\caption{\textbf{Probe Performance Comparison.} The Supervised Ensemble (Idea 4) extracts
some signal, but Consistency (Behavioral) remains superior.}
\label{tab:probe_comparison}
\begin{tabular}{llcc}
\toprule
\textbf{Method} & \textbf{Type} & \textbf{AUROC} & \textbf{Cost (Inference)} \\
\midrule
Random Baseline & Statistical & 0.500 & 1x \\
Focus Entropy ($H_s$) & Unsupervised Visual & 0.504 & 1x \\
Cluster Count ($C_k$) & Unsupervised Visual & 0.501 & 1x \\
\midrule
Linear Probe ($h_{last}$) & Supervised Ling. & 0.620 & 1x \\
\textbf{Ensemble Probe} & \textbf{Supervised Attn.} & \textbf{0.725} & \textbf{1x} \\
\midrule
Self-Consistency (SC) & Behavioral & \textbf{0.784} & 10x \\
\bottomrule
\end{tabular}
\end{table}

\subsection{The Counting Anomaly: Severe Miscalibration}
\label{app:counting}

A critical discovery in our baseline testing was the model's behavior on quantitative
reasoning tasks. We refer to this as the ``Counting Anomaly.''

\textbf{The Phenomenon:}
On tasks asking ``How many [objects] are in the image?'', the evaluated VLM families exhibit
\textbf{severe miscalibration}. As shown in our data, the model often assigns extremely
high probability ($>90\%$) to incorrect integers.

\textbf{Case Study:}
Consider an image with 3 baseball players.
\begin{itemize}
    \item \textbf{Ground Truth:} 3
    \item \textbf{Model Prediction:} ``Four''
    \item \textbf{Token Confidence ($P_{tok}$):} 92\% (Very High)
    \item \textbf{Total Visual Clusters ($K_{\mathrm{total}}$):} 3 distinct clusters (equivalently $C_k=2$ after removing the dominant component).
\end{itemize}

This dissociation highlights a ``Symbolic Detachment.'' The visual encoder correctly
identifies 3 regions (verified by $K_{\mathrm{total}}=3$, hence $C_k=2$), but the projection into the language space
maps these features to the token ``Four.'' Because the language model is autoregressively
coherent, it assigns high probability to the token ``Four'' despite being factually
grounded in ``Three'' visual features.

\textit{Conclusion:} Token probability measures the model's \textit{fluency}, not its
\textit{grounding}. Self-Consistency mitigates this because, in the miscalibrated state,
the model is likely to oscillate between ``Four'' and ``Three'' across different
sampling temperatures, lowering the SC score.

\subsection{Architectural Drivers of Early Locking: Residual Update Analysis}
\label{app:residual_updates}

To investigate the architectural drivers behind LLaVA's ``Early Locking'' and ``Symbolic Detachment'' discussed in Section~\ref{subsec:detachment}, we extracted the hidden states of the 576 visual tokens at every layer of the LLaVA-1.5-7B architecture. We then computed the average $L_2$ norm of the residual updates ($\| h^{(l)} - h^{(l-1)} \|_2$) to measure how actively the model processes visual features at each depth.

\begin{figure}[htbp]
    \centering
    \includegraphics[width=0.7\textwidth]{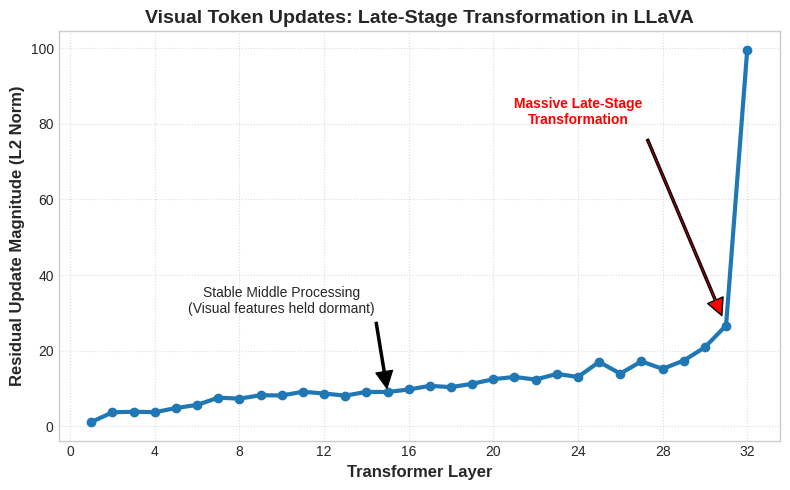}
    \caption{\textbf{Visual Token Updates: Late-Stage Transformation in LLaVA.} We plot the average $L_2$ norm of the residual updates ($\| h^{(l)} - h^{(l-1)} \|_2$) for the 576 visual tokens across all 32 transformer layers. The representations remain largely dormant across the middle layers (Layers 5--28), explaining the stagnation of early attention maps. A massive non-linear transformation occurs only in the final layers (Layers 30--32), forcing the alignment between visual perception and linguistic output.}
    \label{fig:late_stage_forcing}
\end{figure}

As shown in Figure~\ref{fig:late_stage_forcing}, the visual token representations remain remarkably dormant across the middle 25 layers of the network. Because the visual representations are not actively updated during these middle layers, the spatial attention maps naturally stagnate (the ``Early Locking'' phenomenon). The model applies massive non-linear transformations to these features only in the final three layers to extract confidence and generate text, directly corroborating our Logit Lens findings that true visual-linguistic grounding occurs at the end of the network.

\subsection{Qualitative Failure Analysis}
\label{app:qualitative}

We analyzed specific instances where the ``Attention-Confidence Assumption'' broke down.

\textbf{False Negatives (Good Attention, Bad Answer):}
In 15\% of failure cases, the attention map was ``perfect'' (low entropy, high clustering
on relevant objects). For example, in a POPE object-existence query, the model
attended solely to a chair while answering ``No'' to ``Is there a chair?''. This
suggests that the attention mechanism acted as a retrieval query that successfully found
the feature, but the LLM decoder failed to interpret the retrieved feature as ``existence.''

\textbf{False Positives (Bad Attention, Good Answer):}
In 22\% of correct cases, the model exhibited ``scattered'' attention (high entropy,
$H_s > 4.5$). This frequently occurred in background scene questions (e.g., ``Is this
a rainy day?''). The model likely relied on global texture features pooled from the
entire image rather than specific object attention, yet standard interpretability
metrics would penalize this as ``unfocused.''

\subsection{Extended Case Study: Why Attention Fails and Consistency Succeeds}
\label{app:case_study}

\begin{figure}[t]
\centering
\includegraphics[width=0.75\textwidth]{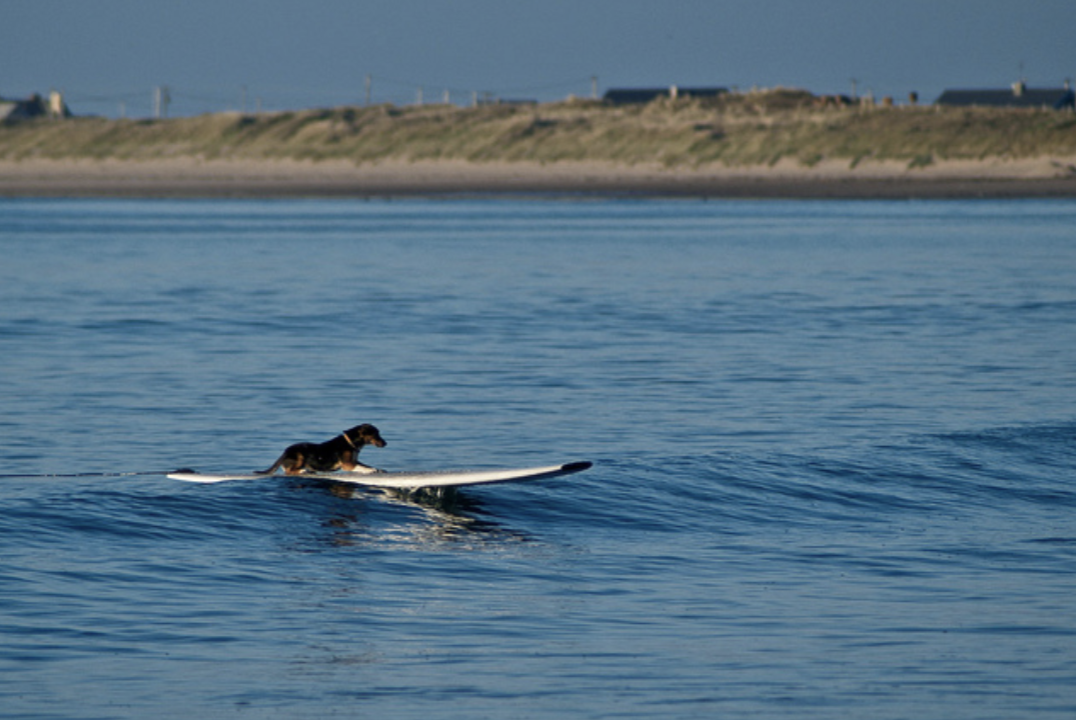}

\vspace{0.5em}
\small
\fbox{\parbox{0.92\textwidth}{
\textbf{Question:} ``Is the dog wearing a collar?'' \hfill \textbf{Ground Truth:} \textcolor{green!60!black}{\textbf{Yes}}

\vspace{0.4em}
\begin{tabular}{p{5.5cm}|p{5.5cm}}
\multicolumn{1}{c|}{\textcolor{blue}{\textbf{Attention Metrics}}} & \multicolumn{1}{c}{\textcolor{green!60!black}{\textbf{Mechanistic Analysis}}} \\
\hline
\\[-0.8em]
Spatial Entropy: $H_s = 0.321$ {\scriptsize (Very low)} & Peak layer: L14 ($\Delta$margin $= +9.57$) \\
Cluster Count: $C_k = 0$ {\scriptsize (Single dominant focus)} & Token ``Yes'' suppressed at L10--14 \\[0.3em]
\textcolor{red}{\ding{55} \textbf{Would predict: Reliable}} & \textcolor{green!60!black}{\ding{51} \textbf{Correctly flags: Unreliable}} \\
\end{tabular}

\vspace{0.4em}
\hrule
\vspace{0.3em}
\centering
\textbf{Model Output:} ``\textcolor{red}{No}'' \quad \textbf{Confidence:} $P = 54.6\%$ \quad \textcolor{red}{\textbf{INCORRECT}}
}}
\caption{\textbf{Case Study: High-Quality Attention, Wrong Answer (PaliGemma, Sample \#31).} The image shows a dog on a surfboard clearly wearing a red collar. The model answers ``No'' despite exhibiting \textit{excellent} attention: very low entropy ($H_s=0.321$, bottom 15\% of dataset) and a single dominant focus ($C_k=0$ under our connected-component definition in Appendix~\ref{app:exp_details}). Attention-based metrics would classify this as trustworthy. However, the logit lens reveals that the correct token ``Yes'' is suppressed at layer 14, correctly identifying unreliability.}
\label{fig:case_study}
\end{figure}

To concretely illustrate the disconnect between visual attention and reliability, we present an actual failure case from our VQAv2 experiments (Figure~\ref{fig:case_study}).

\textbf{Why Attention Fails:} This example starkly illustrates the ``Cluster Failure.'' The model's attention exhibits \textit{ideal} structural properties: entropy $H_s=0.321$ places it in the bottom 15\% (highly focused), and a single dominant focus ($C_k=0$ under our definition) suggests the model is ``looking'' at a specific region. By all attention-based metrics, this should be a reliable prediction. Yet the model hallucinates the absence of a collar that is clearly visible. The failure occurs because attention captures \textit{where} features were extracted, not whether those features were correctly interpreted. The visual encoder successfully attends to the dog, but the downstream LLM fails to bind the ``collar'' concept to the perceived visual features.

\textbf{Why Logit Lens Succeeds:} Probing the hidden states reveals the failure mechanism. The correct token ``Yes'' gains probability through layers 0--10 as visual features are processed, but is sharply suppressed at layer 14: the peak visual integration point ($\Delta$margin $= +9.57$). This suppression pattern, detectable by our hidden-state probes, correctly flags the prediction as unreliable. The model's internal trajectory reveals uncertainty that the final output masks.

This case exemplifies our core finding: \textit{looking well is not knowing well}. A model can attend perfectly to the right region and still hallucinate.

\subsection{Cross-Model Experiment Details}

\subsection{Family-Specific Reliability Patterns}
\label{app:family_patterns}

To complement the per-model deep dives, we summarize family-specific behaviors: \textbf{LLaVA-1.5} shows a long early-lock plateau followed by late diffusion and strong final-layer probe signal; \textbf{PaliGemma-3B} shows earlier integration and weaker late-layer margin separation; and \textbf{Qwen2-VL-7B} shows iterative re-integration cycles with strong late reliability separation.

\label{app:crossmodel}

We conducted extensive experiments across three VLM architectures to validate generality.

\textbf{Model Architectures:}
\begin{itemize}
    \item \textbf{LLaVA-1.5-7B}: 32 transformer layers, 32 attention heads per layer.
    Uses frozen CLIP ViT-L/14 visual encoder with Vicuna-7B language backbone.
    Visual tokens projected via 2-layer MLP.
    \item \textbf{PaliGemma-3B} (Google): 18 transformer layers, 8 attention heads per
    layer. Uses SigLIP visual encoder with Gemma language backbone. Visual tokens
    projected via linear layer.
    \item \textbf{Qwen2-VL-7B-Instruct} (Alibaba): 28 transformer layers with Grouped
    Query Attention (28 heads, 4 KV heads). Native multimodal architecture with
    interleaved visual tokens and dynamic resolution support.
\end{itemize}

\subsection{Model-Specific Deep Dive: LLaVA}
\label{app:llava_detailed}

\textbf{Key Insight:} Correctness emerges \textit{before} final answer selection.
Margin trajectories diverge at Layer 21 and peak at Layer 24, suggesting reliability
is determined in mid-layers, not at the final output. In our notation, this corresponds to $l_{\mathrm{vis}}^{\star}=24$, while the maximum absolute final margin occurs at $l_{\mathrm{final}}^{\star}=31$. Table \ref{tab:llava_complete}
presents the complete LLaVA analysis.

\begin{table}[ht]
\centering
\caption{\textbf{Model-Specific Complete Analysis (LLaVA-1.5-7B).} Layer-wise computational pipeline,
neuron-level findings, and causal validation.}
\label{tab:llava_complete}
\small
\begin{tabular}{llll}
\toprule
\multicolumn{4}{l}{\textit{Layer-wise Computational Pipeline}} \\
\textbf{Layers} & \textbf{Role} & \textbf{$\Delta$margin} & \textbf{Dominant Component} \\
\midrule
0--16 & Feature extraction & Low variance & N/A \\
17 & Early prediction & N/A & 82.3\% probe accuracy \\
19 & Early boosting & +0.53 & MLP \\
21--28 & Suppression & $-$0.85 to $-$2.27 & Attention (72\%) \\
24 & Max separation & N/A & Largest correct/incorrect gap \\
29 & Neuron commitment & N/A & 86.3\% probe, 5.7\% sparse \\
30 & Answer boosting & +2.61 & MLP \\
31 & Final decision & \textbf{+9.20} & MLP (72\%) \\
\midrule
\multicolumn{4}{l}{\textit{Key Neurons (Layer 31)}} \\
\textbf{Neuron ID} & \textbf{Type} & \textbf{$\Delta$activation} & \textbf{Functional Role} \\
\midrule
1512 & Success & +27.23 & Answer confidence \\
1360 & Failure & $-$3.11 & Failure detection \\
3839 & Failure & $-$3.08 & Failure detection \\
2660 & Failure & $-$2.95 & Failure detection \\
\bottomrule
\end{tabular}
\end{table}

\subsection{Implementation and Hardware Details}
\label{app:implementation}

All experiments were conducted on compute clusters provided by RunPod and Lambda Labs, using NVIDIA A100 GPUs (80GB VRAM), AMD EPYC 7742 64-Core CPUs, and 512 GB system memory. The software stack used PyTorch 2.1.0 with CUDA 12.1 and the HuggingFace \texttt{transformers} library with official checkpoints for LLaVA, PaliGemma, and Qwen2-VL \cite{liu2023llava,beyer2024paligemma,wang2024qwen2vl}. Attention extraction was implemented via PyTorch \texttt{register\_forward\_hook} hooks on decoder \texttt{MultiheadAttention} modules in each family's multimodal-integration regime (e.g., late layers for LLaVA and architecture-adjusted regions for PaliGemma and Qwen2-VL).

\subsection{Discussion Extensions: Cross-Family Interpretation and Efficiency Trade-offs}
\label{app:discussion_extra}

\subsection{Cross-Family Interpretation}

Across all three families, the same reliability taxonomy appears with model-specific signatures. \textbf{LLaVA-1.5} exhibits the strongest symbolic-detachment gap (early lock, late diffusion), which aligns with high probe separability in late layers. \textbf{PaliGemma-3B} integrates visual evidence earlier and more smoothly, yielding weaker late-layer separability and lower probe AUROC (0.738). \textbf{Qwen2-VL-7B} shows cyclical refinement and strong late-stage re-separation, consistent with high probe AUROC (0.971).

These differences suggest that reliability probing should be architecturally adaptive (e.g., layer selection and probe capacity per family), rather than assuming a one-size-fits-all late-layer template.

\subsection{Reliability vs. Efficiency Trade-offs}

While Self-Consistency (SC) is the gold standard for reliability (\textbf{$R=0.43$}), it comes
at a high computational cost: it requires $K=10$ forward passes. For real-time
applications (e.g., robotics), this is often prohibitive.

Our \textbf{Hidden State Probe} offers a compelling alternative:
\begin{itemize}
    \item \textbf{Self-Consistency:} High Accuracy ($AUROC=0.78$), High Cost ($10\times$ inference).
    \item \textbf{Learned Probe:} Moderate to High Accuracy (up to $AUROC=0.96$ on family-specific splits),
    Zero Cost (overhead of a single linear layer).
    \item \textbf{Visual Metrics:} Low Accuracy ($AUROC=0.50$), Low Cost.
\end{itemize}

The success of the Hidden State Probe confirms that the model's reliability is encoded
in the \textit{linear subspace} of the final residual stream. This aligns with recent
work in ``Lie Detection'' for LLMs, extending it to the multimodal domain. Future work
should focus on distilling the signal from Self-Consistency into a single-pass value
head, effectively training the model to predict its own consistency score.

\subsection{Limitations and Future Work}
\label{app:limitations}

\textbf{Model Scale:} Our study focuses on three mid-scale open VLMs. It is possible that
larger models (e.g., LLaVA-34B or GPT-4V) exhibit stronger alignment between attention
and truthfulness due to better reinforcement learning from human feedback (RLHF).

\textbf{Computational Cost:} The most reliable metric found, Self-Consistency, requires
$K=10$ inference passes. This is prohibitively expensive for low-latency edge applications.

\textbf{Causal Evidence Scope:} While our ablation experiments demonstrate causal effects of probe-identified neurons (8.3\% accuracy drop for top-5 vs. 0\% for random), the effect requires ablating multiple neurons simultaneously, suggesting a localized circuit rather than individual ``truth units.'' The effect is also moderate in magnitude, indicating these neurons are \textit{contributors} to reliability rather than sole determinants. Future work should explore activation patching and interchange interventions to further characterize the causal mechanism.

\textbf{Future Direction:}
We propose that future work should focus on \textit{distillation}. Since Self-Consistency
provides a high-quality ``silver label'' for reliability ($R=0.43$), we can curate a
dataset of (Image, Question, Answer, SC-Score) and fine-tune a value head on top of
the VLM to predict the SC-Score in a single pass. This would combine the accuracy of
consistency with the efficiency of a probe.

\end{document}